\documentclass[manuscript]{acmart}

\AtBeginDocument{%
  \providecommand\BibTeX{{%
    \normalfont B\kern-0.5em{\scshape i\kern-0.25em b}\kern-0.8em\TeX}}}

\setcopyright{acmcopyright}
\copyrightyear{2018}
\acmYear{2018}
\acmDOI{10.1145/1122445.1122456}

\acmConference[Woodstock '18]{Woodstock '18: ACM Symposium on Neural
  Gaze Detection}{June 03--05, 2018}{Woodstock, NY}
\acmBooktitle{Woodstock '18: ACM Symposium on Neural Gaze Detection,
  June 03--05, 2018, Woodstock, NY}
\acmPrice{15.00}
\acmISBN{978-1-4503-XXXX-X/18/06}

\begin{document}

\title[You Do Not Need a Bigger Boat]{\textit{You Do Not Need a Bigger Boat}\\Recommendations at Reasonable Scale in a (Mostly) Serverless and Open Stack}

\author{Jacopo Tagliabue}
\email{jtagliabue@coveo.com}
\affiliation{%
  \institution{Coveo Labs}
  \city{New York}
  \state{NY}
  \country{USA}
}

\renewcommand{\shortauthors}{Tagliabue}

\begin{abstract}
  We argue that immature data pipelines are preventing a large portion of industry practitioners from leveraging the latest research on recommender systems. We propose our \textit{template} data stack for machine learning at ``reasonable scale'', and show how many challenges are solved by embracing a serverless paradigm. Leveraging our experience, we detail how modern open source can provide a pipeline processing terabytes of data with limited infrastructure work.\footnote{Our title is obviously a small tribute to \textit{Jaws}: \url{https://www.youtube.com/watch?v=z1TzKbF7KyE}.}
\end{abstract}



\maketitle

\section{Introduction}
With almost 4 trillion dollars spent yearly in online retail~\cite{emarketer2020}, research in the eCommerce space gained considerable traction in the last few years, with important insights for recommendation systems~\cite{10.1145/3383313.3412248}, IR/NLP~\cite{CoveoECNLP22,Bi2020ATE,Tagliabue2020ShoppingIT} and more~\cite{Tsagkias2020ChallengesAR}. A quick look at eCommerce workshops at major ML venues reveals an unsettling pattern\footnote{See \textit{Appendix A} for details.}: contributions (and implied benefits) are all but evenly distributed in the market, as the majority of innovation is concentrated in few large players. 

The barrier to entry for cutting-edge recommendation systems in eCommerce is indeed high and multi-faceted: lack of open, representative datasets (as highlighted for example in \cite{CoveoSIGIR2021}), non-relevant benchmarks in the literature (see for example the arguments in~\cite{Requena2020} when replicating~\cite{rakuten2017}), expensive computational resources~\cite{strubell-etal-2019-energy,10.1145/3442188.3445922}. Even when things are smooth on the modelling side, bringing a recommender into production remains a formidable challenge for shops in the mid-long tail, lacking best practices and a tool-chain that works at ``reasonable scale''. In \textit{this} contribution, we tackle this problem directly; in particular: 

\begin{itemize}
    \item we highlight the peculiar constraints (and the opportunities) that lie at ``reasonable scale'' -- that is, mid-to-large shops, with dozens (not hundreds) of ML engineer, making between 10 and 500 million (not billion) USD / year, and producing terabytes (not petabytes) / year of data in behavioral signals;
    \item we show a deep-dive into an end-to-end stack (mostly) built with open-source tools, and show how to productionize a recommender system with (almost) no explicit infrastructure work; we motivate our choices with insights gained by deploying models for dozens of digital shops at all scales.
\end{itemize}

With the growing number of providers in the MLOps space and an ever-changing landscape, a major obstacle to democratization of machine learning is knowing how the tools in the ecosystem play together: the sheer amount of choices to be made may feel overwhelming and the fear of making mistakes may further slow down the adoption of the most appropriate tools. By providing a worked-out example for a recommender pipeline, we hope to provide both a review of important concepts for `reasonable scale'' recommenders, and actionable insights for all the practitioners outside of few retail giants that need to make adoption choices with limited resources\footnote{We will also provide a full implementation as part of the open source project started with \textit{Metaflow}: \url{https://github.com/jacopotagliabue/metaflow-intent-prediction}.}.

\section{Principles for models at reasonable scale}
\label{sec:principles}

Practitioners building models for shops in the mid-long tail face many challenges, as companies which are late adopters of machine learning tools tend to be less mature across the entire stack -- from data collection to model serving. A guiding principle to produce impact quickly and reliably is \textit{independence}: the more data scientists need to rely on other teams (to get data, provision GPUs, serve models, etc.), the more likely something will get ``lost in translation'', and time-to-ROI will soar. On the other hand, we should not assume that data scientists come with an unreasonably complete skill-set: if now it is \textit{their} job to provision GPU, we just shifted the burden, not increased velocity. The following principles condense what we learned by working with dozens of organizations, and provide a framework to make strategic decisions and prioritize resources in a constrained environment:

\begin{enumerate}
    \item \textit{data is king}: the biggest marginal gain is \textit{always} in the data -- making clean and standardized data accessible is significant more important than small modelling adjustments. Indeed, as the market is quickly recognizing, modelling \textit{per se} is getting increasingly commoditized, making proprietary data flows even more important from a strategic point of of view; 
    \item \textit{ELT is better than ETL}: a clear separation between data ingestion and processing produces reliable, reproducible data pipelines. In particular, great care needs to be taken to ingest and store data as an immutable raw record of the state of the system at a given point in time; 
    \item \textit{Paas/Faas is better than Iaas}: maintaining and scaling infrastructure with devoted people is costly -- and unnecessary. At reasonable scale, many providers offer fully-managed services to run our computation without worrying about downtime~\cite{jiang2021towards}, replication, auto-scaling. When resources are constrained, we should invest our time and effort in our core problem -- providing good recommendations --, and buy everything else: we keep the team small and our costs more predictable.\footnote{As predicting the cost of scaling a PaaS service to more users is significantly easier than predict the impact of new hires maintaining a Kubernetes cluster.} The key observation here is that high-quality engineering is the scarcest resource, so we should do everything in our power to have that resource focused on ML. 
    \item \textit{distributed computing is the root of all evil}: distributed systems like Spark played a pivotal role in the Big Data revolution. However, even as a managed service, distributed computing is slow, hard to debug, and force programming patterns unfamiliar to many scientists. At reasonable scale there are better tools to do the heavy lifting for our computations, freeing us completely from distributed computing and all its overhead. 
\end{enumerate}

The key take-away is that working at ``reasonable scale'' comes some advantages: the scale make a lot of tools affordable, and streamline a lot of the complexities needed with more sophisticated systems. As we shall see, by selecting the right tools we can empower relatively small teams to produce a great impact.

\section{Desiderata for In-Session Recommendations}
\label{sec:desiderata}

We use in-session recommendation as an example, and outline what is needed at a functional level for a system to work, from data ingestion to model serving; we then dive deep on how to build a tool chain satisfying these \textit{desiderata}. We chose in-session recommendation as it is a well-studied research topic and a prominent use case for digital shops~\cite{BianchiSIGIReCom2020}:

\begin{itemize}
    \item \textit{raw data ingestion}, which includes getting shopper's data in a scalable way, storing them safely, guaranteeing re-playability from raw events;
    \item \textit{data preparation}, which includes data visualizations and BI dashboard, data quality checks, data wrangling and feature preparation;
    \item \textit{model training}, which includes model training, hyperparameter search, behavioral checklists;
    \item \textit{model serving}, which includes serving predictions at scale;
    \item \textit{orchestration}: which includes a monitoring UI, automated retries, and a notification system.
\end{itemize}

\begin{figure}\centering
    \includegraphics[width=9cm]{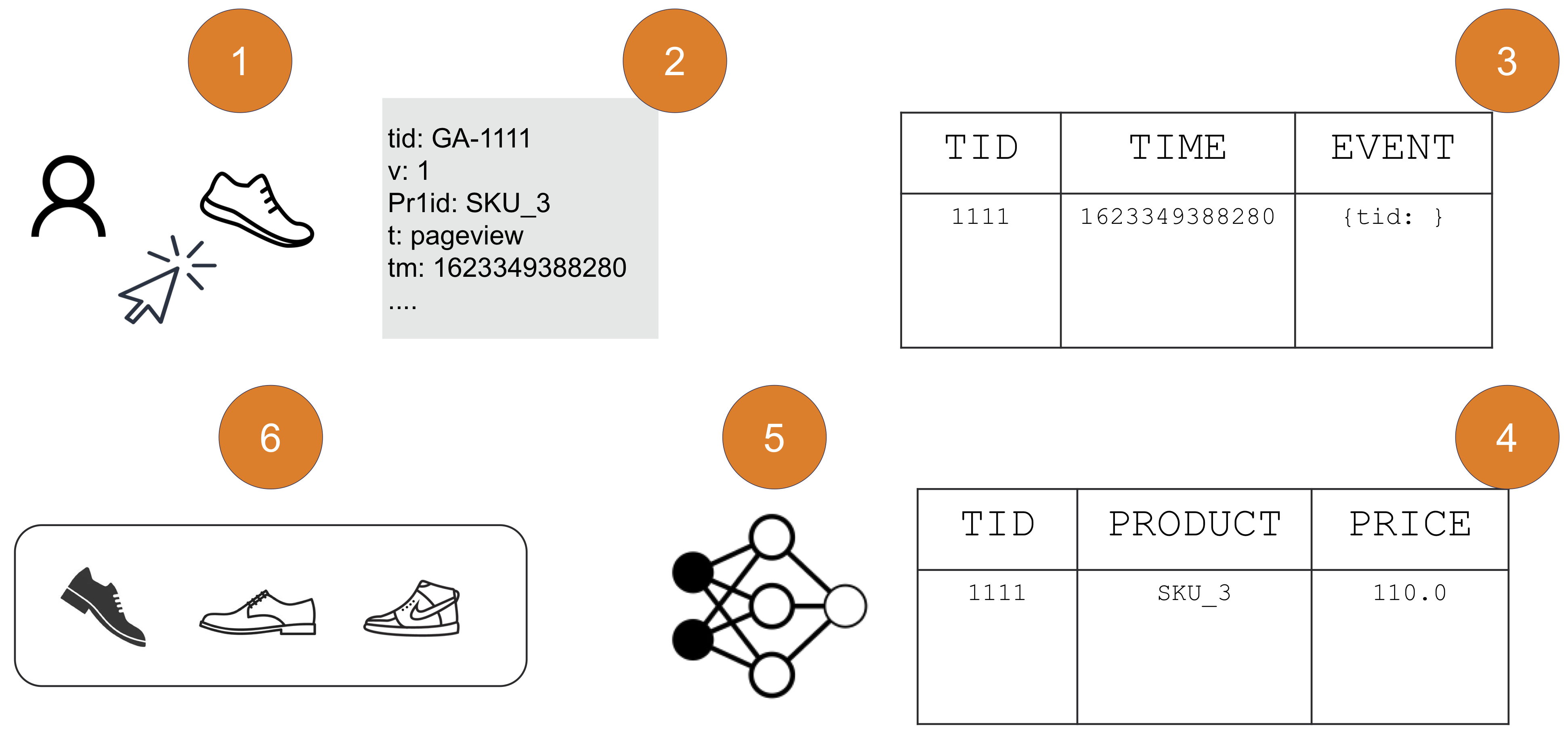} 
    \caption{Data processing from events collected on the browser (shopper clicking on a product in \textbf{1}), to a carousel of recommendations served on the PDP, \textbf{6}. Raw data (\textbf{2)} is sent to a table (\textbf{3}), and stored raw in an append-only fashion. From there, data is transformed in usable format for training (for example, exploding important properties in a devoted table, \textbf{4}), and a model is then trained (\textbf{5}) to serve recommendations (\textbf{6}).}
    \label{fig:pipeline}
\end{figure}

A useful exercise is to visualize the process, and see the journey of a shopping event from the browser (collected with a standard Javascript SDK\footnote{For example, Google Analytics: \url{https://analytics.google.com/analytics/web/}.}), up to the training loop in our machine learning model (see Fig.~\ref{fig:pipeline}).

\section{An End-to-End Stack}

\begin{figure}\centering
    \includegraphics[width=10cm]{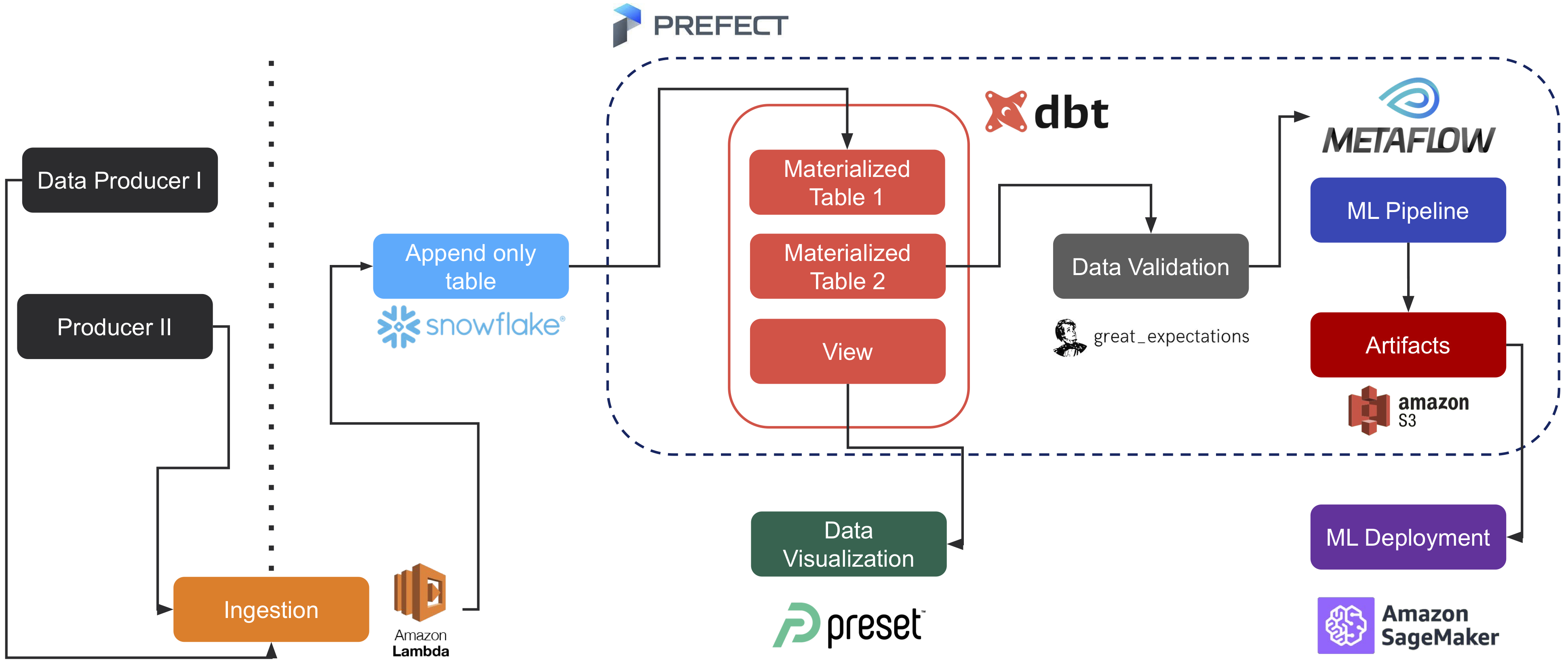} 
    \caption{An end-to-end data stack for companies at ``reasonable scale'', from data ingestion and storage, to visualize, QA, training and serving.}
    \label{fig:stack}
\end{figure}

Fig.~\ref{fig:stack} depicts a modern data stack that combines the principles at ``reasonable scale'' (Section~\ref{sec:principles}) with the functional components from Section~\ref{sec:desiderata}:

\begin{itemize}
    \item \textit{raw data ingestion} is achieved in PaaS (with auto-scaling) through AWS Lambda~\cite{medium2019};\footnote{Please note that the above example feature AWS components, but equivalent modules are available in all major cloud providers.} storage is again achieved in a PaaS-like manner through \textit{Snoflake} \cite{10.1145/2882903.2903741};
    \item \textit{data preparation} starts with \textit{dbt}\footnote{Open sourced at \url{https://www.getdbt.com/}.}, which builds a SQL-based DAG of transformations to prepare normalized tables for data visualization\footnote{\textit{Preset} is a PaaS version of the open-source \textit{Superset}, \url{https://superset.apache.org/}.} and QA\footnote{{Great Expectations} is an open-source tool for data validation, available at \url{https://greatexpectations.io/}.}; 
    \item \textit{model training} happens with \textit{Metaflow}\footnote{Open sourced at \url{https://metaflow.org/}.}), which allows the definition of ML tasks as a DAG, and abstracts away cloud execution (including GPU provisioning) through simple decorators;
    \item \textit{model serving} happens through AWS Sagemaker\footnote{\url{https://aws.amazon.com/it/sagemaker/}.}, which is our hosted tool of choice for serving models in auto-scaling and with a variety of hardware options: please note that since \textit{Metaflow} comes with artefacts and versioning, deployment options are plenty and easy to change;
    \item \textit{orchestration} happens with \textit{Prefect}\footnote{Open sourced at \url{https://www.prefect.io/}.}, which also offers a hosted version for job monitoring and admin purposes.
\end{itemize}

There are three crucial observations of how it all fits together: first, there are no resources directly maintained by ML engineers\footnote{A Prefect agent would be the exception, but that could also be avoided by running on AWS step functions directly).} as all tools are maintained and scaled automatically\footnote{Models inside \textit{Metaflow} may need to still be manually updated, of course, but that is core ML engineering.}; second, the distributed nature of ``reasonable size'' computing is abstracted away in \textit{Snowflake} through plain SQL: everything downstream of data aggregation/preparation can happen comfortably locally; third, warehouse aside, most tools are either already open source, or substitutable with open ones \footnote{Serverless computing for example is available as open-source as well: \url{https://openwhisk.apache.org/}.}.

\section{Conclusion}

We argued that infrastructure and architectural barriers preventing practitioners from leveraging the latest ML research can be surmounted by embracing a serverless paradigm. We know from experience that the stack we proposed (or a similar one) can process terabytes of data (from raw events to GPU-powered recommendations), with limited-to-none devOps work, and mostly relying on a thriving community of open-source solutions. Of course, data and model work~\cite{10.1145/3411764.3445518} still needs to happen, but that is why we built everything in the first place: we should be happy that catching our prey in this growing ecosystem won't require a bigger boat.  

\bibliographystyle{ACM-Reference-Format}
\bibliography{sample-base}

\appendix

\section{APPENDIX: RESEARCH DISTRIBUTION}
\label{app:distribution}
\begin{figure}\centering
    \includegraphics[width=8cm]{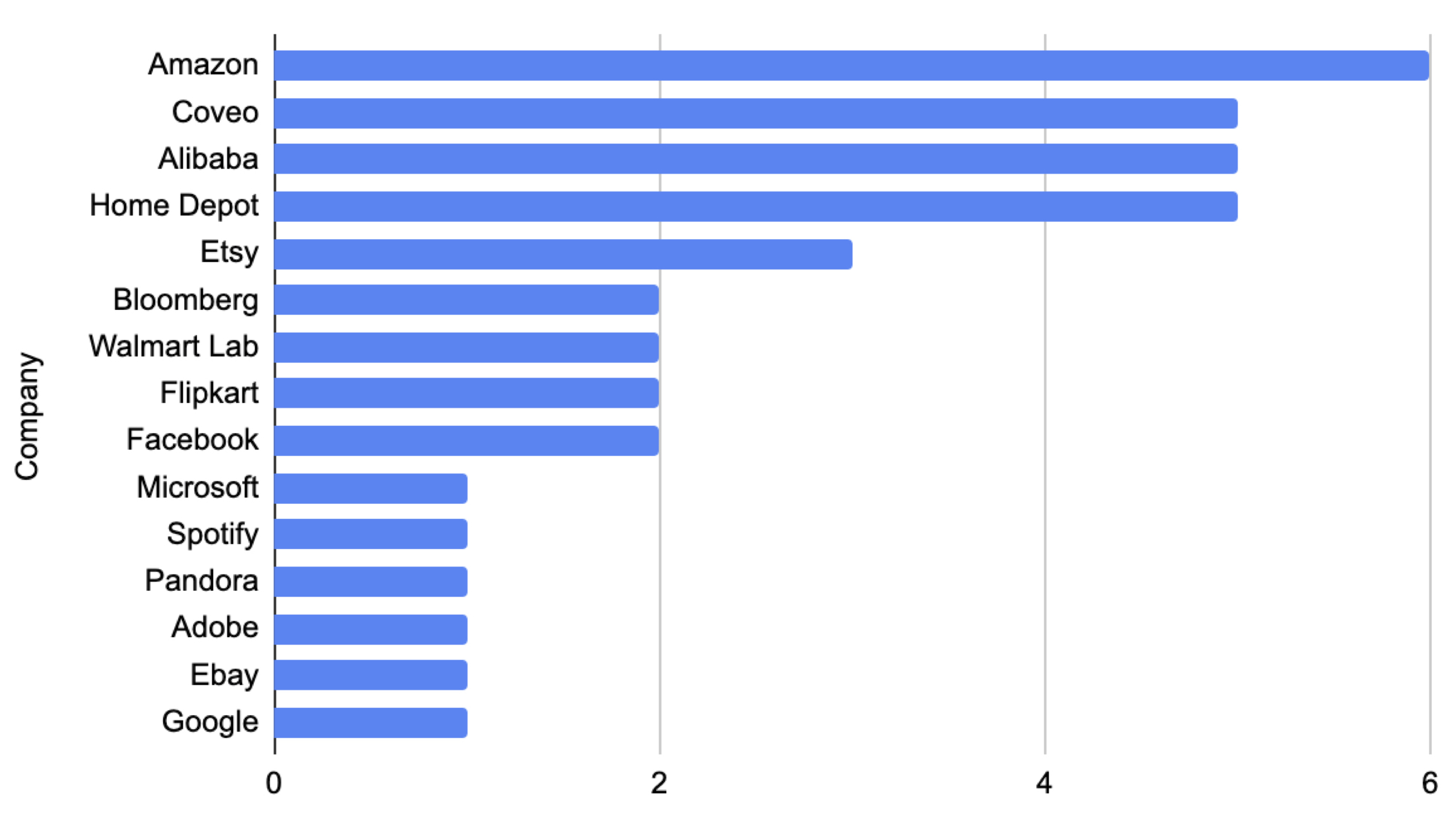} 
    \caption{Number of research papers in eCommerce events at top tier conferences in 2020: almost all contributions are from public companies with a B2C business model.}
    \label{fig:ecommerce}
\end{figure}

Figure~\ref{fig:ecommerce} shows the number of paper per company at eCommerce-themed events at major conferences in 2020 (KDD, SIGIR, ACL, WWW, RecSys). A total of 28 industry players took part in those events; out of 28, only 2 companies are not large public companies, and only one contributed multiple times (\textit{Coveo}, with 6 research papers).

\section{APPENDIX: BIO}
\textit{Jacopo Tagliabue} was co-founder and CTO of Tooso, an A.I. company in San Francisco acquired by Coveo in 2019. Jacopo is currently the Lead A.I. Scientist at Coveo, where he ships machine learning models to hundreds of companies and millions of shoppers. When not busy building A.I. products, he is exploring research topics at the intersection of language, reasoning and learning: he is a committee member for international NLP/IR workshops, and his work is often featured in the general press and A.I. venues (including SIGIR, RecSys, ACL and best industry paper at NAACL). In previous lives, he managed to get a Ph.D., do scienc-y things for a pro basketball team, and simulate a pre-Columbian civilization.

\end{document}